\documentclass{article}

\usepackage{arxiv}

\usepackage[utf8]{inputenc}
\usepackage[T1]{fontenc}
\usepackage[english]{babel}
\usepackage{csquotes}
\usepackage[final,
            tracking=true,
            kerning=true]{microtype}
\usepackage{xspace}

\usepackage{amssymb}
\usepackage{amsmath}
\usepackage{mathtools}
\usepackage{bbm}
\newtheorem{definition}{Definition}
\newtheorem{theorem}{Theorem}

\usepackage{graphicx}
\usepackage{booktabs}

\usepackage{nth}
\usepackage{siunitx}
\usepackage[dvipsnames]{xcolor}

\newcommand*{\titletext}{Adversarial Robustness Curves}

\usepackage[backend=biber,
            sorting=none,
            maxcitenames=2,
            maxbibnames=8]{biblatex}
\usepackage{varioref}
\usepackage[hypertexnames=false,
            colorlinks=true,
            linkcolor=MidnightBlue,
            urlcolor=Maroon,
            citecolor=ForestGreen,
            pdfauthor={Jan Philip Göpfert},
            pdftitle={\titletext}]{hyperref}
\usepackage[all]{hypcap}
\usepackage[noabbrev,
            capitalize,
            nameinlink]{cleveref}

\usepackage[backgroundcolor=white,
            linecolor=lightgray,
            bordercolor=white,
            textsize=tiny]{todonotes}

\newcommand*{\iid}{i.\,i.\,d.\@\xspace}
\newcommand*{\wip}{w.\,p.\@\xspace}
\newcommand*{\uar}{u.\,a.\,r.\@\xspace}
\newcommand*{\R}{\mathbb{R}}

\DeclarePairedDelimiterX{\norm}[1]{\lVert}{\rVert}{#1}

\DeclareMathOperator{\sgn}{sign}

\title{\titletext}
\date{}

\usepackage[auth-sc]{authblk}

\author{Christina Göpfert\textsuperscript{*}}
\author{Jan P. Göpfert\textsuperscript{*}}
\author{Barbara Hammer}
\affil{Bielefeld University, Germany}

\newcommand\blfootnote[1]{{%
  \let\thempfn\relax%
  \footnotetext[0]{#1}%
}}


\begin{document}
\maketitle
\begin{abstract}
The existence of adversarial examples has led to considerable uncertainty
regarding the trust one can justifiably put in predictions produced by automated
systems. This uncertainty has, in turn, lead to considerable research effort in
understanding adversarial robustness. In this work, we take first steps towards
separating robustness analysis from the choice of robustness threshold and norm.
We propose robustness curves as a more general view of the robustness behavior
of a model and investigate under which circumstances they can qualitatively
depend on the chosen norm.
\end{abstract}
\blfootnote{\textsuperscript{*}equal contribution}
\section{Introduction}
Robustness of machine learning models has recently attracted massive research
interest. This interest is particularly pronounced in the context of deep
learning. On the one hand, this is due to the massive success and widespread
deployment of deep learning. On the other hand, it is due to the intriguing
properties that can be demonstrated for deep learning (although these are not
unique to this setting): the circumstance that deep learning can produce models
that achieve or surpass human-level performance in a wide variety of tasks, but
completely disagree with human judgment after application of imperceptible
perturbations~\cite{Szegedy2014Intriguing}. The ability of a classifier to
maintain its performance under such changes to the input data is commonly
referred to as \emph{robustness to adversarial perturbations}.

In order to better understand adversarial robustness, recent years have seen the
development of a host of methods that produce adversarial examples, in the white
box and black box settings, with specific or arbitrary target labels, and
varying additional
constraints~\cite{Goodfellow2014Explaining,Kurakin2016Adversarial,Kurakin2016Adversariala,Papernot2016Practical,Su2017One}.
There has also been a push towards training regimes that produce adversarially
robust networks, such as data augmentation with adversarial examples or
distillation~\cite{Papernot2016Distillation,Gu2014Towards,Huang2015Learning,Bastani2016Measuring}.
The difficulty faced by such approaches is that robustness is difficult to
measure and quantify: even if a model is shown to be robust against current
state of the art attacks, this does not exclude the possibility that newly
devised attacks may be successful~\cite{Carlini2017Towards}. The complexity of
deep learning models and counter-intuitive nature of some phenomena surrounding
adversarial examples further make it challenging to understand the impact of
robust training or the properties that determine whether a model is robust or
non-robust. Recent work has highlighted settings where no model can be
simultaneously accurate and robust~\cite{Tsipras2019Robustness}, or where
finding a model that is simultaneously robust and accurate requires optimizing
over a different hypothesis class than finding one that is simply
accurate~\cite{Nakkiran2019Adversarial}. These examples rely on linear models,
as they are easy for humans to understand. They analyze robustness properties
for a fixed choice of norm and, typically, a fixed disadvantageous perturbation
size (dependent on the model). This raises the question: “How do the presented
results depend on the choice of norm, choice of perturbation size, and choice of
linear classifier as a hypothesis class?”

In this contribution, we:
\begin{itemize}
    \item propose \emph{robustness curves} as a way of better representing
    adversarial robustness in place of “point-wise” measures,
    \item show that linear classifiers are not sufficient to illustrate all
    interesting robustness phenomena, and
    \item investigate how robustness curves may depend on the choice of norm.
\end{itemize}
\section{Definitions}
 In the following, we assume data $(x,y) \in X \times Y$, $X \subseteq \R^d$,
 are generated i.i.d. according to distribution $P$ with marginal $P_X$. Let
 $f:X \to Y$ denote some classifier and let $x \in X$. The \emph{standard loss}
 of $f$ on $P$ is
\begin{equation}
    L(f) := P(\{(x,y) : f(x) \neq y\})\,.
\end{equation}
Let $n:X \to \R^+$ be some norm, let $\varepsilon \geq 0$ and let
\begin{equation}
    B_n(x, \varepsilon) := \{x' : n(x - x') \leq \varepsilon \}\,.
\end{equation}
Following~\cite{Tsipras2019Robustness}, we define the
\emph{$\varepsilon$-adversarial loss} of $f$ regarding $P$ and $n$ as
\begin{equation}
    L_{n,\varepsilon}(f) := P(\underbrace{\{(x,y) : \exists x' \in B_n(x,
    \varepsilon): f(x') \neq y \}}_{=:A_\varepsilon^n}) \,. \label{eq:untargeted
    adv loss}
\end{equation}
We have $L_{n,0}(f) = L(f)$. Alternatively, we can exclude from this definition
any points that are initially misclassified by the model, and instead consider
as adversarial examples all points where the model changes its behavior under
small perturbations.
Then the \emph{$\varepsilon$-margin loss} is defined as
\begin{equation}
    L'_{n,\varepsilon}(f) := P_X(\{x: \exists x' \in B_n(x, \varepsilon): f(x')
    \neq f(x) \})\,. \label{eq:untargeted margin loss}
\end{equation}
$L'_{n,\varepsilon}$ is the weight of all points within an $\varepsilon$-margin
of a decision boundary. We have $L'_{n,0}(f) = 0$.

There are two somewhat arbitrary choices in the definition in
\cref{eq:untargeted adv loss,eq:untargeted margin loss}: the choice of
$\varepsilon$ and the choice of the norm $n$. The aim of this contribution is to
investigate how $\varepsilon$ and $n$ impact the adversarial robustness.
\section{Robustness Curves}
As a first step towards understanding robustness globally, instead of for an
isolated perturbation size $\varepsilon$, we propose to view robustness as a
function of $\varepsilon$. This yields an easy-to-understand visual
representation of adversarial robustness in the form of a \emph{robustness
curve}.
\begin{definition}
	The \emph{robustness curve} of a classifier $f$, given a norm $n$ and
	underlying distribution $P$, is the curve defined by
	\begin{align}
	r_{f,n,P} : [0, \infty) &\to [0,1] \\
	\varepsilon &\mapsto L_{n, \varepsilon}(f) \,.
	\end{align}
	The \emph{margin curve} of $f$ given $n$ and $P$ is the curve defined by
	\begin{align}
	r'_{f,n,P} : [0, \infty) &\to [0,1] \\
	\varepsilon &\mapsto L'_{n, \varepsilon}(f) \,.
	\end{align}
\end{definition}
Commonly chosen norms for the investigation of adversarial robustness are the
$\ell_1$ norm (denoted by $\|\cdot\|_1$), the $\ell_2$ norm (denoted by
$\|\cdot\|_2$), and the $\ell_\infty$ norm (denoted by $\|\cdot\|_\infty$). In
the following, we will investigate robustness curves for these three choices of
$n$.

\begin{figure}
    \centering
    \includegraphics[width=5in]{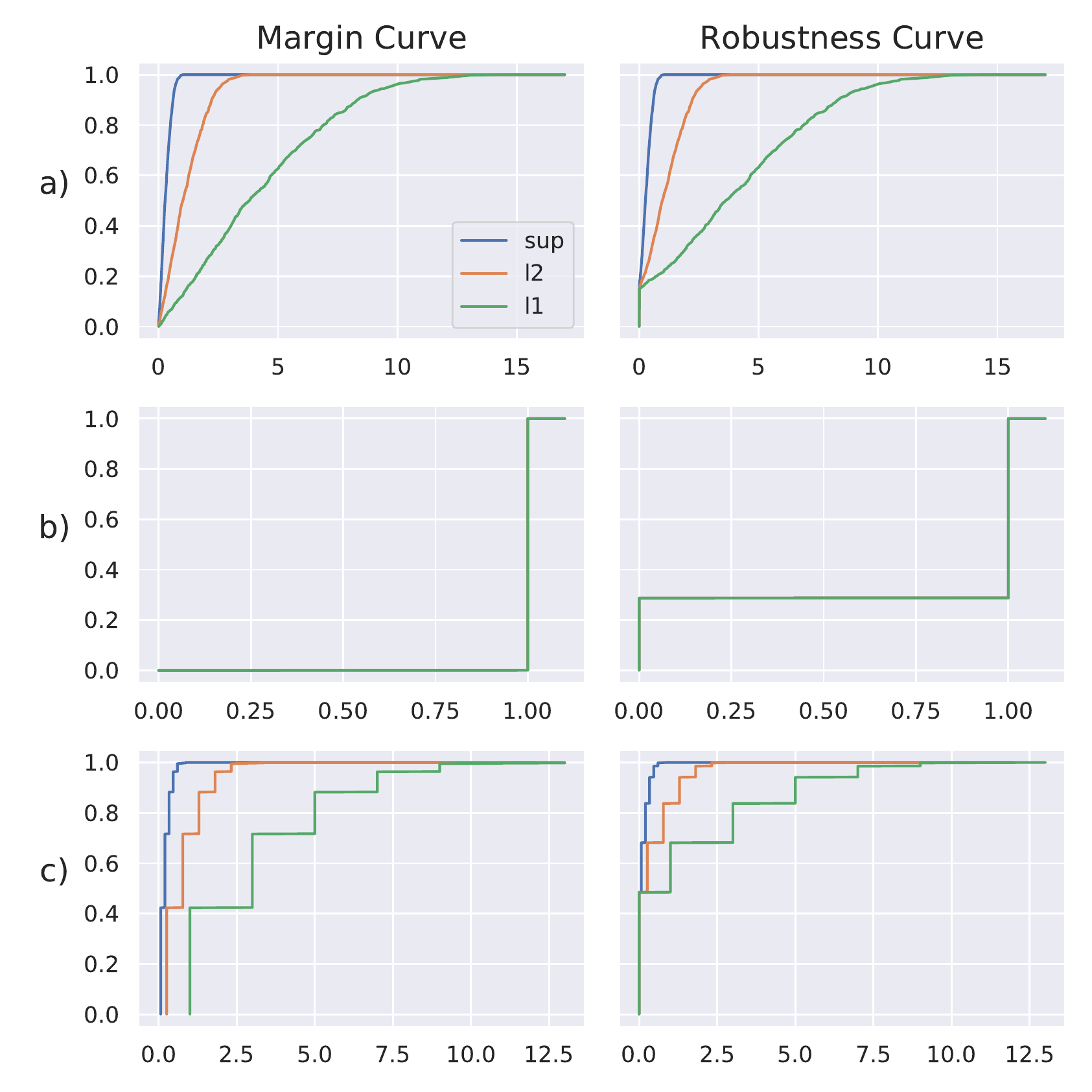}
    \caption{Margin curves and robustness curves for several examples of
    distributions and linear models from the literature. Row a) shows curves for
    classifier $f_{\mathrm{avg}}$ and distribution $P_1$. Row b) shows curves
    for classifier $f_{\mathrm{rob}}$ and distribution $P_1$. In this case, all
    three curves are identical and thus appear as one. Row c) shows curves for
    classifier $f_s$ and distribution $P_2$.}
    \label{fig:lin_curves}
\end{figure}

\Textcite{Tsipras2019Robustness} propose a distribution $P_1$ where $y
\stackrel{\text{\uar}}{\sim} \{-1, +1\}$ and
\begin{equation}
x_1 = \begin{cases} 1 & \text{\wip} \, p \\ -1 & \text{\wip} \, (1-p)
\end{cases} \quad x_2, \dots, x_{d+1} \stackrel{\text{\iid}}{\sim}
\mathcal{N}(\eta y, 1).
\end{equation}
For this distribution, they show that the linear classifier $f_{\mathrm{avg}}(x)
= \sgn(w^T x)$ with $w = (0, 1/d, \dots, 1/d)$ has high accuracy, but low
$\varepsilon$-robustness in $\ell_\infty$ norm for $\varepsilon \geq 2 \eta$,
while the classifier $f_{\mathrm{rob}}(x) = \sgn(w^Tx)$ with $w = (1, 0, \dots,
0)$ has high $\varepsilon$-robustness for $\varepsilon < 1$, but low accuracy.
\Textcite{Nakkiran2019Adversarial} proposes a distribution $P_2$ where $y
\stackrel{\text{\uar}}{\sim} \{-1, +1\}$ and
\begin{equation}
x_i = \begin{cases} y & \text{\wip} \, 0.51 \\ -y & \text{\wip} \, 0.49 \end{cases}
\end{equation}
where the linear classifier $f_{s}(x) = \sgn(w^T x)$ with $w = \vec{1}_d$ has
high accuracy, but low $\varepsilon$-robustness in $\ell_\infty$ norm for
$\varepsilon \geq \frac{1}{2}$. \Cref{fig:lin_curves} shows margin curves and
robustness curves for $P_1$ and $f_{\mathrm{avg}}$, $P_1$ and $f_{\mathrm{rob}}$
and $P_2$ and $f_s$.
\section{The impact of \texorpdfstring{$n$}{n}}
The curves shown in \cref{fig:lin_curves} seem to behave similarly for each
norm. Is this always the case? Indeed, if $f$ is a linear classifier
parameterized by normal vector $w$ and offset $b$, denote by
\begin{equation}
    d_n((w, b),  x) = \min \{n(v):  \exists p: x = p + v, \langle  w , p \rangle + b = 0\} 
\end{equation}
the shortest distance between $(w, b)$ and $ x$ in norm $n$. Then a series of
algebraic manipulations yield
\begin{align}
    d_{\| \cdot \|_1}((w,b), x) &=  \frac{|b + \langle w, x \rangle|}{\|w\|_\infty} \,,\\
    d_{\| \cdot \|_2}((w,b), x) &=  \frac{|b + \langle w, x \rangle|}{\|w\|_2} \,, \\
    d_{\| \cdot \|_\infty}((w,b), x) &= \frac{|b + \langle w, x \rangle|}{\|w\|_1}\,.
\end{align}
In particular, there exist constants $c$ and $c'$ depending on $(w,b)$ such that
for all $x \in X$,
\begin{equation}
    d_{\| \cdot \|_1}((w,b), x) = c d_{\| \cdot \|_2}((w,b), x) = c' d_{\| \cdot \|_\infty}((w,b), x)
\end{equation}
This implies the following Theorem:
\begin{theorem}
For any linear classifier $f$, there exist constants $c, c' > 0$ such that for
any $\varepsilon \geq 0$,
\begin{equation}
L_{\|\cdot\|_1, \varepsilon}(f) = L_{\|\cdot\|_2, \varepsilon / c}(f) =  L_{\|\cdot\|_\infty, \varepsilon / c'}(f)\,.
\end{equation}
\end{theorem}
As a consequence, for linear classifiers, dependence of robustness curves on the
choice of norm is purely a matter of compression and elongation.

What can we say about classifiers with more complex decision boundaries? For all
$x$, we have
\begin{equation}
\|x\|_\infty \leq \|x\|_2 \leq \|x\|_1 \leq \sqrt{d} \|x\|_2 \leq d \|x\|_\infty\,. \label{eq:norm relationship}
\end{equation}
These inequalities are tight, i.e. there for each inequality there exists some
$x$ such that equality holds. It follows that, for any $\varepsilon > 0$,
\begin{align}
A_{\varepsilon/d}^{\|\cdot\|_\infty} \subseteq A_{\varepsilon/ \sqrt{d}}^{ \|\cdot\|_2} \subseteq A_\varepsilon^{\|\cdot\|_1} \subseteq A_{\varepsilon}^{\|\cdot\|_2} \subseteq A_{\varepsilon}^{\|\cdot\|_\infty}
\end{align}
and so
\begin{align}
L_{\|\cdot\|_\infty, \varepsilon}(f) &\geq L_{\|\cdot\|_2, \varepsilon}(f) \geq L_{\|\cdot\|_1, \varepsilon}(f) \\
&\geq L_{\|\cdot\|_2, \varepsilon / \sqrt{d}}(f) \geq L_{\|\cdot\|_1, \varepsilon/d}(f) \,.
\end{align}
In particular, the robustness curve for the $\ell_\infty$-norm is always an
upper bound for the robustness curve for any other $\ell_p$-norm (since $\|x\|_p
\leq \|x\|_\infty$ for all $x$ and $p \geq 1$). Thus, for linear classifiers as
well as classifiers with more complicated decision boundaries, in order to show
that a model is adversarially robust for any fixed norm, it is sufficient to
show that it exhibits the desired robustness behavior for the
$\ell_\infty$-norm. On the other hand, in order to show that a model is
\emph{not} adversarially robust, showing this for the $\ell_\infty$ norm does
not necessarily imply the same qualities in another norm, as the robustness
curves may be strongly separated in high-dimensional spaces, both for linear and
non-linear models.

\begin{figure}
    \centering
    \includegraphics[width=5in]{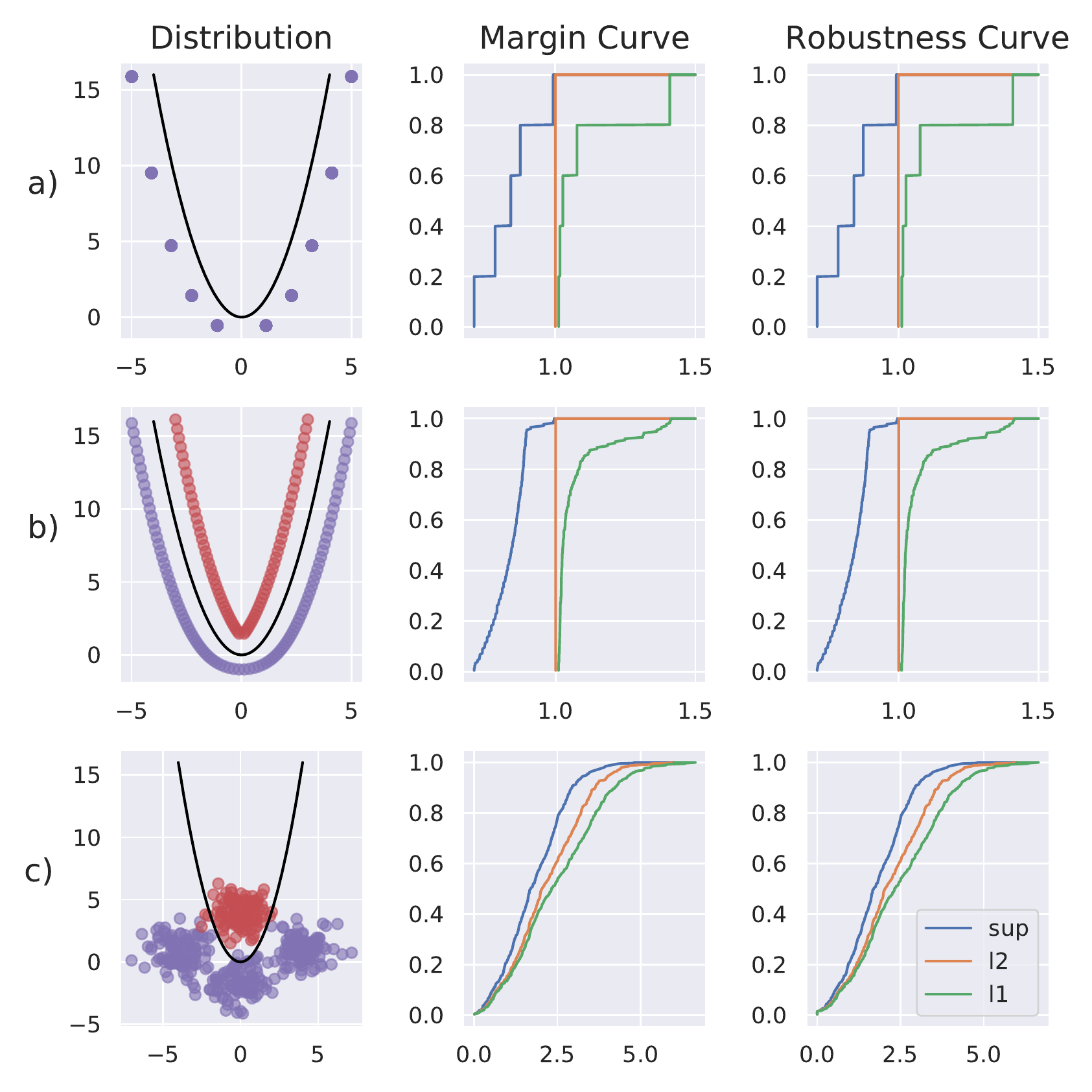}
    \caption{Margin curves and robustness curves for $f(\vec x) = \sgn(x_1^2 -
    x_2)$ and three different underlying distributions, illustrating varying
    behavior of the robustness curves for different norms. In rows a) and b),
    the robustness curves are qualitatively different, while they are almost
    identical in row c). Note that in these examples, robustness curves and
    margin curves are nearly identical, as the standard loss of $f$ is zero or
    close to zero in all cases.}
    \label{fig:parabola_curves}
\end{figure}

Contrary to linear models, for more complicated decision boundaries, robustness
curves may also exhibit \emph{qualitatively} different behavior. This is
illustrated in \cref{fig:parabola_curves}. The decision boundary in each case is
given by a quadratic model in 2-dimensional space: $f(\vec x) = \sgn(x_1^2 -
x_2)$. In the first example, we construct a finite set of points, all at
$\ell_2$-distance 1 from the decision boundary, but at various $\ell_1$ and
$\ell_\infty$ distances. For any distribution concentrated on a set of such
points, the $\ell_2$-robustness curve jumps from zero to one at a single
threshold value, while the $\ell_1$- and $\ell_\infty$-robustness curves are
step functions with the height of the steps determined by the distribution
across the points and the width determined by the variation in $\ell_1$ or
$\ell_\infty$ distances from the decision boundary. The robustness curves in
this example also exhibit, at some points, the maximal possible separation by a
factor of $\sqrt{d}$ (note that $d=2$) while touching in other points. In the
second example, we show a continuous version of the same phenomenon, with points
inside and outside the parabola distributed at constant $\ell_2$-distance from
the decision boundary, but with varying $\ell_1$ and $\ell_\infty$ distances. As
a result, the robustness curves for different norms are qualitatively different.
The third example, on the other hand, shows a setting where the robustness
curves for the three norms are both quantitatively and qualitatively similar.

These examples drive home two points:
\begin{itemize}
	\item The robustness properties of a classifier may depend both
	quantitatively and qualitatively on the norm chosen to measure said
	robustness. When investigating robustness, it is therefore imperative to
	consider which norm, or, more broadly, which concept of closeness best
	represents the type of perturbation to guard against.
	\item Linear classifiers are not a sufficient tool for understanding
	adversarial robustness in general, as they in effect neutralize a degree of
	freedom given by the choice of norm.
\end{itemize}
\section{Discussion}
We have proposed robustness curves as a more general perspective on the
robustness properties of a classifier and have discussed how these curves can or
cannot be affected by the choice of norm. Robustness curves are a tool for a
more principled investigation of adversarial robustness, while their dependence
on a chosen norm underscores the necessity of basing robustness analyses on a
clear problem definition that specifies \emph{what kind} of perturbations a
model should be robust to. We note that the use of $\ell_p$ norms in current
research is frequently meant only as an approximation of a “human perception
distance”~\cite{HiddenInPlainSight}. A human's ability to detect a perturbation
depends on the point the perturbation is applied to, meaning that human
perception distance is not a homogeneous metric, and thus not induced by a norm.
In this sense, where adversarial robustness is meant to describe how faithful
the behavior of a model matches that of a human, the adversarial loss in
\cref{eq:untargeted adv loss} can only be seen as a starting point of analysis.
Nonetheless, since perturbations with small $\ell_p$-norm are frequently
imperceptible to humans, adversarial robustness regarding some $\ell_p$-norm is
a reasonable lower bound for adversarial robustness in human perception
distance. In future work, we would like to investigate how robustness curves can
be estimated for deep networks and extend the definition to robustness against
targeted attacks.
\AtNextBibliography{\raggedright\small}
\printbibliography
\end{document}